\documentclass[10pt,twocolumn,letterpaper]{article}

\usepackage{iccv}
\usepackage{times}
\usepackage{epsfig}
\usepackage{graphicx}
\usepackage{amsmath}
\usepackage{amssymb}

\usepackage[breaklinks=true,bookmarks=false]{hyperref}

\iccvfinalcopy 

\def\iccvPaperID{****} 

\ificcvfinal\fi

\begin{document}

\title{Aerial multi-object tracking by detection using deep association networks}

\author{Ajit Jadhav\\
Indian Institute of Information Technology, Sri
City\\
{\tt\small jadhavajit.j16@iiits.in}
\and 
Prerana Mukherjee\\
Indian Institute of Technology, Sri
City\\
{\tt\small prerana.m@iiits.in}
\and
Vinay Kaushik\\
Indian Institute of Technology, Delhi\\
{\tt\small vinaykaushik15@gmail.com}
\and
Brejesh Lall\\
Indian Institute of Technology, Delhi\\
{\tt\small brejesh@ee.iitd.ac.in}
}

\maketitle
\ificcvfinal\thispagestyle{empty}\fi

\begin{abstract}
   A lot a research is focused on object detection and it  has achieved significant advances with deep learning techniques in recent years. Inspite of the existing research, these algorithms are not usually optimal for dealing with sequences or images captured by drone-based platforms, due to various challenges such as view point change, scales, density of object distribution and occlusion.
In this paper, we develop a model for detection of objects in drone images using the VisDrone2019 DET dataset. Using the RetinaNet model as our base, we modify the anchor scales to better handle the detection of dense distribution and small size of the objects. We explicitly model the channel interdependencies by using “Squeeze-and-Excitation” (SE) blocks that adaptively recalibrates channel-wise feature responses. This helps to bring significant improvements in performance at a slight additional computational cost. Using this architecture for object detection, we build a custom DeepSORT network for object detection on the VisDrone2019 MOT dataset by training a custom Deep Association network for the algorithm.

\end{abstract}

\section{Introduction}
Object detection and tracking has remained an important research problem in computer vision \cite{dalal2005histograms, felzenszwalb2009object, liu2016ssd, zhang2018single}. It is relevant for myriad of applications such as video surveillance, scene understanding, semantic segmentation, object localization, robot manipulation etc. In real time scenarios, object detection can pose several challenges such as scale, pose, illumination variations, occlusion, clutter etc. In case of videos, the additional challenge is due to the motion information in dynamic environments. We deal with a specialized category of drone images where the major challenge is posed due to fine granularity and absence of strong discriminative features to handle the inter and intra class variance. In case of unmanned aerial vehicles (UAVs), for autonomous navigation identification of obstacles for a height is very relevant. Drones are generally used for patroling border areas which cannot be done by manual military forces. The typical application ranges from tracking criminals in surveillance videos \cite{wang2013intelligent}, search and rescue \cite{yuan2016statistical}, sports analysis and scene understanding \cite{yuan2017hdpa, meng2017seeds, huang2010biologically, wu2018toward}. There are certain other challenges which are specific to drone images such as density of objects is huge, smaller scale, camera motion constraints and realtime deployment issues. Motivated by these issues, we focus on object detection and tracking in aerial imagery.

\begin{figure*}
    \centering
    \includegraphics[scale=0.7]{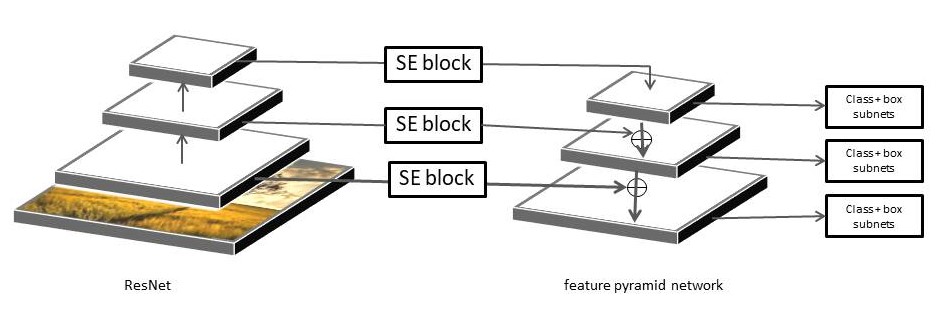}
    \caption{Detection Network}
    \label{fig:my_label1}
\end{figure*}

Owing to the flexibility of drone usage and navigation capabilities, the acquired images can also be utilized to perform 3D reconstruction and object discovery. However, in order to do so techniques resorting to simultaneous localization and mapping (SLAM) based algorithms are required which are again heavily dependent on several other sensor based data such as accelerometer, gyroscope, magnetometer etc. Further, the task of objection detection or collision avoidance methods typically require huge computational overhead. In case of mobile drone videos, the deep learning techniques require to process the images in real time with high accuracy rates. There are two most popularly used frameworks for object detection: i) two-stage framework and ii)
single-stage framework. The two-stage framework represented by R-CNN \cite{girshick2014rich} and its variants \cite{girshick2015fast, ren2015faster, dai2016r, DBLP:journals/corr/abs-1711-07264, DBLP:journals/corr/abs-1906-09756} extract object proposals followed by object classification and bounding box regression. The single stage framework, such as YOLO \cite{redmon2016you, redmon2017yolo9000, redmon2018} and SSD \cite{liu2016ssd,fu2017dssd}, apply object classifiers and bounding box regressors in an end-to-end manner without explicitly extracting object proposals. Most of the state-of-the-art methods \cite{redmon2016you, ren2015faster,  redmon2017yolo9000, redmon2018, DBLP:journals/corr/abs-1808-01244, DBLP:journals/corr/abs-1804-06215, li2019intelligent} typically focus on detecting generic objects from natural images, where most of the targets are sparsely distributed with fewer numbers. However, due to the intrinsic data distribution differences between drone images and natural images, the traditional CNN-based methods tend to miss such densely distributed small objects.



In this paper, we provide a novel multi-object tracking by detection framework particularly for aerial images captured by drones. We detect ten predefined categories of objects (i.e., pedestrian, person, car, van, bus, truck, motor, bicycle, awning-tricycle, and tricycle) in drone images collected for VisDrone 2019 dataset \cite{zhu2019visdrone}. In view of above discussions, the key contributions can be summarized as follows,
\begin{itemize}
    \item We utilize denser anchor scales with large scale variance to detect the dense distribution of smaller objects.
    \item We utilize Squeeze-and-Excitation (SE) \cite{hu2018squeeze} blocks to capture the channel dependencies which results in better feature representation for the detection task in moving camera constraints.
    \item For the tracking model, we train the deep association network \cite{wojke2018deep} on the object hypotheses generated from the detection module and feed it to the the deep sort algorithm \cite{wojke2017simple} for tracking.
\end{itemize}

Remaining sections in the paper are organized as follows.
In Sec. \ref{sec:related} we discuss the related work in object detection
and tracking. In Sec. \ref{sec:methodology}, we outline the methodology
we propose to detect objects and subsequently track them. In Sec. \ref{sec:results}, we
discuss experimental results and conclude the paper in Sec.
\ref{sec:conclusion}.

\section{Related Work}
\label{sec:related}
In this section we provide a detailed overview of the contemporary
techniques prevalent in the domains which are
closely related in this context.
\subsection{Aerial imagery object detection}
In \cite{zhu2019visdrone},release a challenge dataset over drone images with varying weather and lighting conditions. A thorough review of the latest techniques on the benchmark dataset is provided with exhaustive evaluation protocols. In \cite{li2019intelligent}, the authors utilize novel real time object detection and tracking deep learning based algorithms over mobile devices with drones. In \cite{mitrokhin2018event}, authors present object detection method for data collected with asynchronous drone cameras. Spatio-temporal information is captured from the event stream after motion compensation is applied for object localization in motion. In \cite{boudjit2015detection}, authors provide an autonomous target detection and tracking algorithm for AR Parrot Drone. In \cite{huang2017redbee}, authors provide real-time motion detection algorithm for visual inertial drone systems in case of dynamic backgrounds. This can run on low-power application Snapdragon processor with efficient performance capabilities. In \cite{coluccia2017drone}, authors release an interesting challenge dataset for bird vs drone detection in order to prevent smuggling using drones in shore areas. The idea is to generate an alert in case of presence of drones in videos where there might be birds as well flying in the air. In \cite{yanmaz2018drone}, authors provide an architecture for collaborative aerial system with autonomous networking capabilities in aerial traffic. It consists of multi-drone systems consisting of quadcoptors fitted with various on-board sensing devices for communication. It aids in several applications such as disaster assistance, aerial monitoring and search and rescue operations. In \cite{aker2017using}, authors provide an end-to-end trainable deep architecture for drone detection by leveraging data augmentation techniques. In \cite{hsieh2017drone}, authors propose novel Layer Proposal Networks for localizing and counting the number of objects in a dynamic environment. They leverage the spatial layout information in the kernels for improving the localization accuracy.

\subsection{Multi-object tracking}
In \cite{fang2018recurrent}, authors propose a temporal generative network namely recurrent autoregressive network to model the appearance and motion features in temporal sequences. It strongly couples internal and external memory with the network thus incorporating information about previous frames trajectories and long term dependencies. In \cite{kim2018multi}, in order to efficiently learn the long-term appearance models via a recurrent network, Bilinear LSTM based technique is proposed. In \cite{zhu2018online}, authors utilize the advantages of single object tracking and data association methods to detect and track objects in noisy environments. In \cite{henschel2018fusion}, authors postulate the tracking problem as a weighted graph labeling problem. They fuse the head and full body detectors for tracking purposes. In \cite{yoon2018online}, authors provide mechanisms o handle temporal errors in tracking such as drifting and track ID switches. This happens due to occlusion or noise present in the scene. Thus, they incorporate motion and shape information in a siamese network to improve tracking performance. In \cite{zhou2018deep}, authors propose Deep Continuous Conditional Random Field (DCCRF) for handling inter-object relation and movement patterns in tracking. In \cite{ovsep2018track}, authors introduce a category agnostic detection free tracker using segmentation masks with semantic segmentation based approaches. In \cite{beard2018solution}, authors propose a generalised labeled multi-Bernoulli (GLMB) filter for large scale multi-object tracking.  

\subsection{Motion segmentation}
Unsupervised motion segmentation is very important task leveraging object localization as well as adaptive video compression. In \cite{keuper2018motion}, authors provide motion segmentation and tracking by co-clustering techniques. Motion segmentation is performed by grouping of the trajectories. In \cite{ranjan2019competitive}, authors provide a joint framework for unsupervised learning of depth, motion and optical flow to perform motion segmentation by exploiting geometric constraints. In \cite{hu2018unsupervised}, authors adopt saliency estimation with spatial neighborhood information in a graph modeling framework. They utilize optical flow and edge cues for feature extraction. In \cite{mitrokhin2019ev}, authors introduce motion event dataset. They utilize Structure from Motion (SfM) based pipeline with computationally efficient deep neural network for event detection. They rely on dense depth map computation for motion segmentation and estimate the 3D velocities for moving objects. 

\section{Methodology}
\label{sec:methodology}
The VisDrone dataset comprises of images taken at varying altitudes and egocentric movements due to high-altitude wind speeds leading to drastic scale change and occlusions in the scene. The Detection and Tracking (DnT) framework is optimized for handling such scenarios. A large fraction of objects are small and dense which generic DnT frameworks are unable to detect which eventually becomes basis of every tracking scheme. A better detection framework not only ensures the detection is good but also provides a good basis for tracking. Since we track using object to object association in sequential frames, need for an optimal detector becomes more significant.
We describe our DnT architecture for object detection and tracking illustrated in Fig. \ref{fig:my_label1}. The first section puts forward in detail, the selection of RetinaNet as the base deep learning architecture for object detection on the drone dataset. We construct a novel training strategy consisting of a combination of optimal set of anchor scales and utilization of SE blocks for detection and learning a deep association network for tracking detected images in the subsequent frames.

\subsection{Selection of Base Detector: YOLOv3 vs RetinaNet}
We evaluate the results of two single-stage object detectors: YOLOv3 and RetinaNet. For the YOLO model, we use the same training parameters as mentioned in \cite{redmon2018yolov3}  and instead of using the original set of variable square input sizes of {320, 352, 384, 416, 448, 480, 512, 544, 576, 608} we use a set of larger input sizes of {544, 576, 608, 640, 672, 704, 736, 768, 800} to account for high scale and variablitiy of the images in the VisDrone dataset. For this algorithm, on the COCO dataset the 9 clusters for anchors were: (10 × 13), (16 × 30), (33 × 23), (30 × 61), (62 × 45), (59 × 119), (116 × 90), (156 × 198), (373 × 326). We use the same clusters for training our model on the VisDrone-DET dataset.
For the RetinaNet network, we use the same parameters for training the model as mentioned in \cite{lin2017focal} while increasing the input size to 1500 × 1000 and increasing the maximum number of detections to 500. We select RetinaNet as our base Detector as it outperforms YOLOv3 on the VisDrone Dataset.

\begin{figure*}
    \centering
    \includegraphics[scale=0.7]{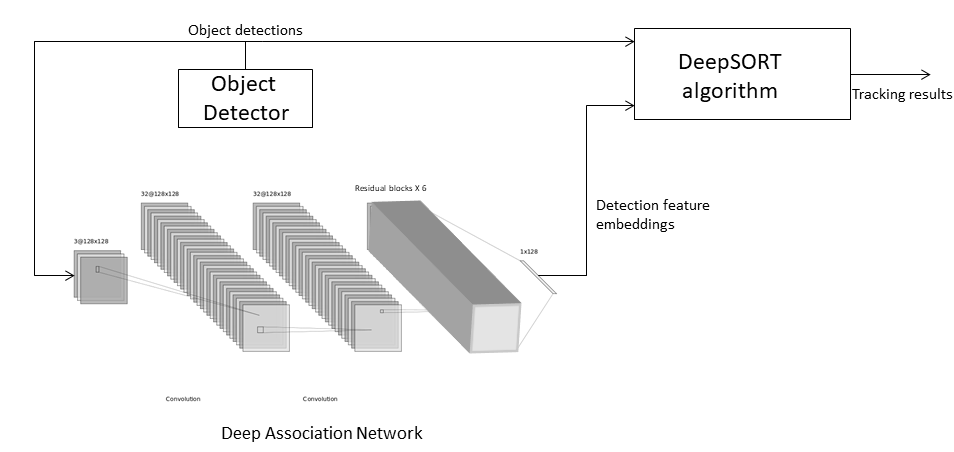}
    \caption{Tracking Network}
    \label{fig:my_label2}
\end{figure*}{}

\subsection{Anchor scales}
One of the most important design factors in a one-stage detection system is how densely it covers the space of possible image boxes. Thus, the anchor box parameters in RetinaNet \cite{lin2017focal}, are critical in creating a Detection framework that is robust to varying object scales. RetinaNet uses translation-invariant anchor boxes. On pyramid levels P3 to P7 in RetinaNet, the anchors have areas of 32*32 to 512*512. At each pyramid level anchors at three aspect ratios {1:2, 1:1, 2:1} are used and anchors of sizes {20, 21/3, 22/3} of the original set of 3 aspect ratio anchors are used for denser scale coverage, at each level. In total there are A = 9 anchors per level and across levels they cover the scale range 32 -813 pixels with respect to the network’s input image.
The anchor parameters used for the original RetinaNet architecture are suited for object detection on natural images. However, as a large number of objects in the VisDrone2019 dataset have a size smaller than 32*32 pixels, many of them having a size nearly equal to 8*8 pixels,  the standard anchor parameters are not the best fit for detecting objects in drone images. This results in objects which don’t have any anchors assigned to them, resulting in these objects not contributing to the training of the model and thus, the model is unable to identify such small objects.
To address this issue,we modify the anchor parameters to cover the range of sizes of objects in the dataset. While we use the same anchor sizes, anchor aspect ratios and strides for the anchors, we use the scales {0.1, 0.25, 0.5, 1, 21/3, 2.2}, which cover a larger variance in size as well as are denser due the use of 6 scales instead of the original 3. This results in assigning anchors to the smaller sized objects more effectively resulting in them contributing to the training and better training of the model.

\subsection{SE Blocks}


In RetinaNet, we generate the set of feature maps {P3, P4, P5, P6, P7} using the feature activation outputs by each stage’s last residual block for the ResNet backbone architecture. Specifically, we use the output of the last residual blocks {C3, C4, C5} which denote the ouputs of conv3, conv4, conv5. We modify the architecture by using passing the outputs {C3, C4, C5} through a SE block before feeding them to the feature pyramid network. This leads to better represented features  for generation of {P3, P4, P5, P6, P7} resulting in better detection results.

\subsection{Multi-Object Tracking Framework}
A multi-object tracking model is built using the detection model for detecting objects in the frames. Similar to DeepSORT, our algorithm learns a deep association network using patches from COCO dataset which enables us in scoring patches on the basis of deep feature similarity. Unlike DeepSORT, we keep track of identity labels for multiple objects of similar classes. Also, when matching detections from subsequent frames, we associate a confidence measure which is provided by the detector and fuse it with the deep association metric, thereby improving tracking for scenarios where confidence score of detected object in the next frame is high but the deep association is low. 

First, the detections are generated from the frames using the object detection model and then the feature embeddings are generated using the trained Deep Association model. The detections including object labels and confidence scores along with the feature embeddings are then passed to the algorithm similar to DeepSORT, which generates the object tracklets based on the detections. 

\begin{figure*}
    \centering
    \includegraphics[width=160mm,height=50mm]{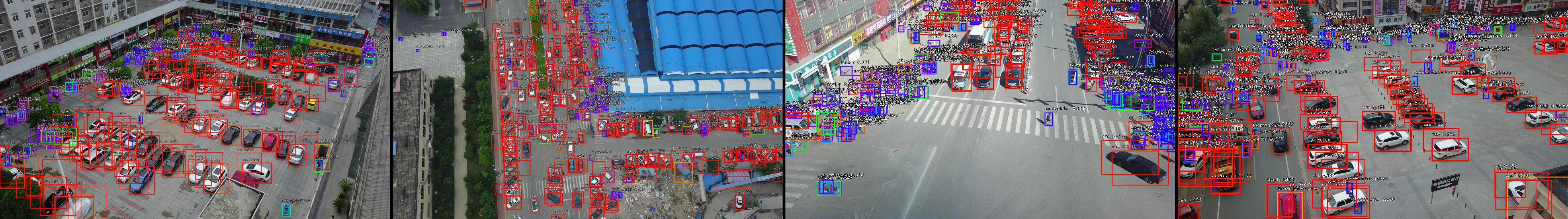}
    \caption{Qualitative Results}
    \label{fig:my_label3}
\end{figure*}

\subsection{Training Strategy}
RetinaNet is trained with stochastic gradient descent (SGD). All models are trained with initial learning rate of 1e-5 with weight decay of 0.0001 and momentum of 0.9 is used. The training loss is the sum the focal loss and the standard smooth L1 loss used for box regression \cite{girshick2015fast}.
To improve speed, we only decode box predictions from at most 1k top-scoring predictions per FPN level, after thresholding detector confidence at 0.05. The top predictions from all levels are merged and class-wise non-maximum suppression with a threshold of 0.5 is applied to yield the final detections. The same parameters mentioned above were used for training all the models.
The base RetinaNet model was trained for 26 epochs with 1618 iterations per epoch using a batch size of 4. The model with improved scales was trained for 25 epochs with 3246 iterations per epoch using a batch size of 4. Finally, the model having new scales along with the SE blocks was trained for 27 epochs with 3246 iterations per epoch using a batch size of 2.

\section{Experimental results and analysis}
\label{sec:results}
The DET framework was evaluated using Visdrone2019 challenge dataset which comprises of multi object detection and tracking datasets. In this section, we describe in detail the optimized hyper-parameters and the intricate implementation details. The proposed DET framework is evaluated on the VisDrone2019 \cite{zhu2019visdrone} dataset benchmarks.

\begin{table}[]
\centering
\begin{tabular}{llll}\hline
Method \textbackslash AP@IoU                                                                                 & 0.50:0.95 & 0.50  & 0.75  \\\hline \hline
Yolo v3                                                                             & 13.8      & 30.43 & 11.18 \\
RetinaNet                                                                           & 14.45     & 23.74 & 15.14 \\
\begin{tabular}[c]{@{}l@{}}RetinaNet\\ (dense scales)\end{tabular}                  & 15.39     & 33.13 & 13.07 \\
\begin{tabular}[c]{@{}l@{}}RetinaNet \\ (dense scales\\ +SE attention)\end{tabular} & 17.19     & 37.69 & 13.97 \\\hline
\end{tabular}
\caption{Average Precision at maxDetections=500}
\label{tab:tab1}
\end{table}

\subsection{Dataset}

VisDrone2019 is a large-scale visual object detection benchmark, which was collected in a very wide area from 14 different cities in China. 
For object detection, it consists of 6,471 images in the training set and 548 images. It has a total of 10 categories, consisting of real-world scenarios such as pedestrian, car, bus, etc. captured using multiple drones with different models under various weather and lighting conditions. 
VisDrone-DET dataset\footnote{It can be downloaded from the following link: http:// www.aiskyeye.com.}, focuses on detecting ten predefined categories of objects (i.e., pedestrian, person, car, van, bus, truck, motor, bicycle, awning-tricycle, and tricycle) in images from drones. Since the dataset consists of default test and train splits, we divide the training set into Train and Validation Splits and select our base network architecture based on the validation results. We finetune our results using the same approach and test on the test set provided in the dataset.

\subsection{Evaluation Metrics}
Output of the algorithm consists of output list of detected bounding boxes with confidence scores for each image. Following the evaluation protocol in MS COCO \cite{lin2014microsoft}, we use the AP IoU=0.50:0.05:0.95 , AP IoU=0.50 , AP IoU=0.75, AR max=1, AR max=10, AR max=100 and AR max=500 metrics to evaluate the results of detection algorithms. These criteria penalize missing detection of objects as well as duplicate detections (two detection results for the same object instance). Specifically, APIoU=0.50:0.05:0.95 is computed by averaging over all 10 Intersection over Union (IoU) thresholds (i.e., in the range [0.50 : 0.95] with the uniform step size 0.05) of all categories, which is used as the primary metric for evaluation and comparison of models. APIoU=0.50 and APIoU=0.75 are computed at the single IoU thresholds 0.5 and 0.75 over all categories, respectively. The ARmax=1, ARmax=10, ARmax=100 and ARmax=500 scores are the maximum recalls given 1, 10, 100 and 500 detections per image, averaged over all categories and IoU thresholds.

\begin{table}[]
\centering
\begin{tabular}{lllll}\hline
Method \textbackslash AR@maxDets                                                                          & 1    & 10   & 100   & 500   \\\hline \hline
Yolo v3                                                                             & 0.36 & 2.63 & 17.53 & 19.34 \\
RetinaNet                                                                           & 0.59 & 5.91 & 20.96 & 21.38 \\
\begin{tabular}[c]{@{}l@{}}RetinaNet\\ (dense scales)\end{tabular}                  & 0.48 & 4.78 & 22.02 & 30.49 \\
\begin{tabular}[c]{@{}l@{}}RetianNet \\ (dense scales\\ +SE attention)\end{tabular} & 0.52 & 4.69 & 23.44 & 31.93\\ \hline
\end{tabular}
\caption{Average Recall at IoU 0.50:0.95}
\label{tab:tab2}
\end{table}

\begin{table*}
\centering
\begin{tabular}{cccccccc}\hline
Method                 & AP[\%] & AP50[\%] & AP75[\%] & AR1[\%] & AR10[\%] & AR100[\%] & AR500[\%] \\\hline \hline
CornerNet\cite{DBLP:journals/corr/abs-1808-01244}              & 17.41 & 34.12   & 15.78   & 0.39   & 3.32    & 24.37    & 26.11    \\
Light-RCNN \cite{DBLP:journals/corr/abs-1711-07264}            & 16.53 & 32.78   & 15.13   & 0.35   & 3.16    & 23.09    & 25.07    \\
DetNet \cite{DBLP:journals/corr/abs-1804-06215}                & 15.26 & 29.23   & 14.34   & 0.26   & 2.57    & 20.87    & 22.28    \\
RefineDet512 \cite{zhang2018single}        & 14.9  & 28.76   & 14.08   & 0.24   & 2.41    & 18.13    & 25.69    \\
Retinanet \cite{li2019intelligent}         & 11.81 & 21.37   & 11.62   & 0.21   & 1.21    & 5.31     & 19.29    \\
FPN  \cite{lin2017feature}             & 16.51 & 32.2    & 14.91   & 0.33   & 3.03    & 20.72    & 24.93    \\
Cascade-RCNN \cite{DBLP:journals/corr/abs-1906-09756}          & 16.09 & 16.09   & 15.01   & 0.28   & 2.79    & 21.37    & 28.43    \\
\textbf{Ours} & 11.19 & 25.65   & 8.78    & 0.56   & 4.87    & 17.19    & 24.09  \\ \hline
\end{tabular}
\caption{Detection Results}
\label{tab:tab3}
\end{table*}

\subsection{Implementation Details}

We use Resnet-50 as the backbone for our detection architecture \cite{he2016deep}. We also use pretrained weights from COCO \cite{lin2014microsoft} dataset for initialization of all our models \cite{deng2009imagenet}. The network architecture is shown in Fig. \ref{fig:my_label2}. In the training stage, the input images are upsampled to 1500 × 1000. For the data augmentation, we use a standard combination of random transform techniques such as rotation, translation, shear, scaling and horizontal flipping. In the test stage, we do not fix the image size and set the confidence threshold to 0.05. We train the network for 50K iterations with the batch size set to 1. The stochastic gradient descent (SGD) solver is adopted to optimize the network with the base learning rate set to 1e-5.

For multi-object tracking, the patches generated from our object detector on MS COCO detection dataset \cite{lin2014microsoft} are resized to 128*128 and fed to the Deep Association network for training. The initial learning was set to 1e-3. The network was regularized with a weight decay of 1 × 10−8 and dropout inside the residual units with probability 0.4. The model was trained for 120k iterations with a batch size of 128.
\begin{table*}
\centering
\begin{tabular}{ccccccccccc}\hline
Method        & AP    & AP@0.25 & AP@0.50 & AP@0.75 & AP car & AP bus & AP truck & AP ped & AP van &      \\ \hline \hline
cem \cite{andriyenko2011multi}          & 5.7   & 9.22    & 4.89    & 2.99    & 6.51   & 10.58  & 8.33     & 0.7    & 2.38   \\
cmot \cite{bae2014robust}        & 14.22 & 22.11   & 14.58   & 5.98    & 27.72  & 17.95  & 7.79     & 9.95   & 7.71   \\
gog  \cite{pirsiavash2011globally}         & 6.16  & 11.03   & 5.3     & 2.14    & 17.05  & 1.8    & 5.67     & 3.7    & 2.55   \\
h2t \cite{wen2014multiple}          & 4.93  & 8.93    & 4.73    & 1.12    & 12.9   & 5.99   & 2.27     & 2.18   & 1.29   \\
ihtls \cite{dicle2013way}        & 4.72  & 8.6     & 4.34    & 1.22    & 12.07  & 2.38   & 5.82     & 1.94   & 1.4    \\
\textbf{Ours} & 13.88 & 23.19   & 12.81   & 5.64    & 32.2   & 8.83   & 6.61     & 18.61  & 3.16  \\ \hline
\end{tabular}
\caption{Tracking Results}
\label{tab:tab4}
\end{table*}
\subsection{Performance Evaluation}

As shown in the results in Table \ref{tab:tab1}, we see that RetinaNet performs better on the VisDrone dataset based on the AP metric where AP score of YOLO is 13.8 while that of RetinaNet is 14.45. Also, we can see that the APIoU=0.5 score is 30.43 while it’s APIoU=0.75 score is 11.18 while for RetianNet APIoU=0.5 score is 23.74 while it’s APIoU=0.75 score is 15.14. The huge drop in the AP value YOLO for higher values of IoU indicates that while it is able to detect objects better than RetinaNet, it struggles to localize the object detections effectively which is an inherent issue with the YOLO architecture. So, we proceed our studies by building a better model based on the RetinaNet architecture. The qualitative results are shown in Fig. \ref{fig:my_label3}.

As shown in Table \ref{tab:tab1}, the initial base RetinaNet model achives an AP score of 14.45 with ARmax=500 score of  21.38\%. For our model with new dense scales we achieve an AP score of 15.39 which is an approximate 6\% increase over the base RetinaNet model. Also, we get an ARmax=500 score of 31.49\% for this model thus, we have a much higher recall due to the increased number of detections as a consequence of using denser scales resulting in better detection of objects across a large variance of object sizes in the dataset. After using SE block along with this architecture, while we only have small increment from 30.49\% to 31.93\% in the  ARmax=500 ,we see a significant ~12\% increase in the AP score to give us an AP score of 17.19. This indicates that while we don’t have significant increase in the number of objects detected, the detected objects are better localized compared to the previous model which results in a higher AP score. This is also proved by APIoU=0.50 and APIoU=0.75 seen in Table \ref{tab:tab1}. where we see that the APIoU=0.50 value increased from 33.13 to 37.69 and the APIoU=0.75 value increased from 13.07 to 13.97. This indicates increase in AP values across all detection thresholds and thus, we can see that the objects are better localized due to the use of better represented features obtained by explicitly modelling interdependencies between channels  by use of SE blocks.

Table \ref{tab:tab2} shows the Average Recall score for different number of maximum detections in the scene on VisDrone detection validation split. Vanilla RetinaNet performs better than standard Yolo v3 on all AR scores. For our model with new dense scales, we achieve better recall rates when the number of detections are high. At maxDets=500, the dense scales model increases the average recall from 21.38\% to 30.49\%. Incorporation of Squeeze and Excitation blocks, further improves the AR for all maxDets especially when the number of detections are greater than 100. The final model increases AR from 30.49\% to 31.93\% for maxDets=500.

Table \ref{tab:tab3} shows VisDrone 2019 detection results evaluated on the provided test set. We can observe that even when our method gives sub-optimal average precision, it performs drastically well for average recall for top 1 and top 10 detections. This has an optimal effect on our tracking pipeline. 
Although the trained Detector performs well on validation set, it performs sub-optimally on the test set. This means possibility of better generalization and more emphasis on smaller objects. The skewness of data is a larger problem that makes learning all the classes difficult. As can be seen from Table \ref{tab:tab4}, our method performs better on smaller objects like pedestrians and cars than all the other methods, and on par with other methods for larger objects such as trucks, vans, buses,etc. 

Also we observe that although the trained detector isn't the most optimal one, our tracker is still able to achieve higher accuracy than almost all the baselines. This proves the robustness of our tracker. Even when the tracked objects have low confidence, the deep association network correctly matches the same object in the subsequent frames. This is due to combined learning of similarity based on deep feature embedding and detection scores. 

\section{Conclusion}
\label{sec:conclusion}
Aerial Object detection problem is an important but preliminary step for the main task of Aerial Multi-Object Tracking. Large number of average confidence detections are preferable than less number of high confidence detections to build an optimal tracker. We presented an efficient tracking and detection framework that performs substantially well on VisDrone DET and MOT datasets respectively. We empirically choose RetinaNet as our base architecture and modify the anchor scale parameter for handling multi-scale dense objects in the scene. We also incorporate SE blocks enabling adaptive re-calibration of channel-wise feature responses. We show that although our method does not achieve overall best results on the detection model, it surpasses other methods as we increase the maximum number of detections. Our tracking pipeline utilizes the same idea and constructs feature embeddings from a trained deep association network along with generated detections and their confidence scores to create labeled tracks for every detected object. It should be emphasied that the proposed framework aims to improve multi-object tracking for aerial imagery. Not surprisingly, the uneven class distribution of data makes it difficult to learn features for all objects which can also be seen in the results. This can be improved in future by better augmentation methods, collecting more relevant data and incorporating structure similarity losses. Similarly, certain conditions like high camera motion, complex motion dynamics, occlusions create problems in tracking. However, these types of situations require a better understanding of the physics of scene such as flow maps, depth maps and semantic maps etc. which is beyond the scope of this paper. 

{\small
\bibliographystyle{ieee}
\bibliography{egbib}
}
\end{document}